\documentclass[conference]{IEEEtran}
\IEEEoverridecommandlockouts

\usepackage[colorlinks=true,allcolors=black]{hyperref} %
\usepackage[all]{hypcap}
\usepackage{cite}
\usepackage{amsmath,amssymb,amsfonts}
\usepackage{algorithmic}
\usepackage{graphicx}
\usepackage{textcomp}
\usepackage{multirow}
\usepackage{comment}
\usepackage{booktabs}
\usepackage{xcolor}
\usepackage{pifont}
\usepackage{tikz}
\renewcommand{\thefigure}{\arabic{figure}}

\def\BibTeX{{\rm B\kern-.05em{\sc i\kern-.025em b}\kern-.08em
    T\kern-.1667em\lower.7ex\hbox{E}\kern-.125emX}}
    
\begin{document}

\title{\Huge Interpretable Logical Anomaly Classification via Constraint Decomposition and \\Instruction Fine-Tuning
}


\author{
\IEEEauthorblockN{Xufei Zhang, Xinjiao Zhou, Ziling Deng, Dongdong Geng, Jianxiong Wang }
 \IEEEauthorblockA{\textit{Beijing XingYun Digital Technology Co., Ltd.}}
}
\maketitle

\begin{abstract}
Logical anomalies are violations of predefined constraints on object quantity, spatial layout, and compositional relationships in industrial images. While prior work largely treats anomaly detection as a binary decision, such formulations cannot indicate which logical rule is broken and therefore offer limited value for quality assurance. We introduce Logical Anomaly Classification (LAC), a task that unifies anomaly detection and fine-grained violation classification in a single inference step. To tackle LAC, we propose \emph{LogiCls}, a vision–language framework that decomposes complex logical constraints into a sequence of verifiable subqueries. We further present a data-centric instruction synthesis pipeline that generates chain-of-thought (CoT) supervision for these subqueries, coupling precise grounding annotations with diverse image-text augmentations to adapt vision language models (VLMs) to logic-sensitive reasoning. Training is stabilized by a difficulty-aware resampling strategy that emphasizes challenging subqueries and long tail constraint types. Extensive experiments demonstrate that \emph{LogiCls} delivers robust, interpretable, and accurate industrial logical anomaly classification, providing both the predicted violation categories and their evidence trails. 
\end{abstract}


\section{Introduction}
\label{sec:intro}

Anomaly detection and classification~\cite{gu2024anomalygpt,li2024lr,zhang2024featureicassp,zhu2025fine} are indispensable for industrial visual inspection, directly affecting yield, rework cost, and downstream process stability. Industrial anomalies can be broadly categorized into structural and logical. Structural anomalies such as cracks, scratches, or contamination typically manifest as localized appearance defects and have been extensively studied with representation learning and reconstruction-based pipelines. Logical anomalies, in contrast, arise from violations of high level constraints on quantity, spatial arrangement, and composition, where evidence is often subtle, globally distributed, and only meaningful under a rule-based interpretation of the scene.

This constraint-centric view also appears in recent controllable generation and editing research, which increasingly emphasizes explicit constraint satisfaction such as consistent object quantity and layout~\cite{shen2025imagharmony}, subject consistent transformations under user intent~\cite{shen2025imagedit}, and fine-grained controllable garment design~\cite{shen2025imaggarment}. Related studies on pose-guided generation further highlight that global structure and spatial configuration must be treated as first class constraints rather than incidental appearance cues~\cite{shen2024imagpose}. These advances suggest that high level constraints are central to modern visual reasoning and controllability. For industrial inspection, the corresponding requirement is to verify such constraints from observations under data scarcity and subtle visual cues, which makes logical anomaly understanding substantially more challenging than structural defect recognition.

\begin{figure}[t]
    \centering
    \includegraphics[width=\linewidth]{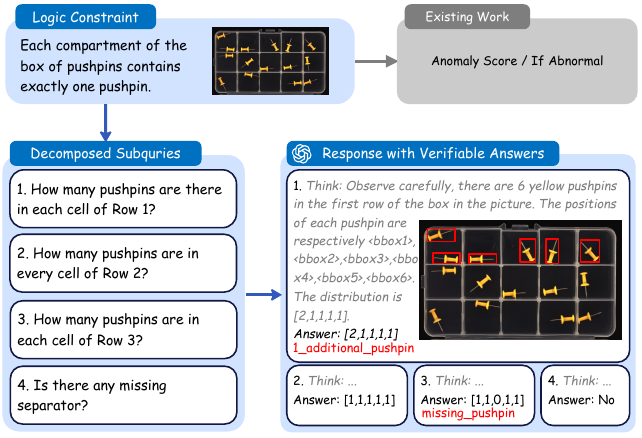}
    \vspace{-0.3cm}
    \caption{Comparison of different approaches to logical anomalies. Existing methods typically output an anomaly score or a normal abnormal binary decision, whereas our setting aims to explicitly identify normal samples and classify specific logical violation types.}
    \label{fig:introduction}
\end{figure}

Research on structural anomaly detection and classification is well established~\cite{batzner2024efficientad,zuo2024reconstructionicassp,lyu2025mvrec,huang2025anomalyncd,cao2024adaclip}. Logical anomalies were introduced as a distinct benchmark direction with MVTec LOCO~\cite{bergmann2022mvtechloco}. While recent work begins to study logical anomaly detection, fine-grained classification of logical violations remains largely underexplored. From a modality perspective, existing approaches fall into two lines. The first relies primarily on the visual modality, which is necessary but often insufficient to capture compositional constraints and long range dependencies. The second leverages multimodal feature matching to model attributes, entity relations, and higher order compositions. More recently, training free systems have emerged that concatenate multiple base models and query powerful vision language models to reach a final decision~\cite{zhang2025logsad,jin2025logicad}. Although effective, such multi-component pipelines can be operationally heavy, sometimes depending on online services, which leads to complex workflows and high deployment costs.

To bridge this gap, we introduce Logical Anomaly Classification, which determines whether an image contains a logical anomaly and identifies its specific violation category. To better reflect real world compositions, we reconstruct a dataset such that the training split contains normal samples and single anomaly samples, while the test split includes images with multiple coexisting logical anomalies. Addressing Logical Anomaly Classification with a compact vision language model presents several challenges. Logical structures are governed by multiple interdependent constraints. Many violations are subtle and sparsely represented in the training split. Small-scale models have limited spatial and numerical reasoning ability, and the dataset scale and diversity are constrained.

We propose \emph{LogiCls}, a framework that couples fine-grained synthetic data construction with instruction fine-tuning to endow compact vision language models with verifiable logical reasoning. The key idea is to decompose complex constraints into atomic and verifiable subqueries grounded in the image, each designed to map precisely to a specific anomaly category. We then construct an instruction dataset that integrates object grounding, chain of thought reasoning, and image text augmentation, encouraging the model to produce structured and interpretable outputs. During training, a difficulty-aware resampling strategy adjusts sampling probabilities based on prediction errors and uncertainty, improving performance on hard cases. At inference, the compact model answers the subqueries and aggregates their results into a final anomaly decision along with transparent evidence trails.

In summary, our contributions are:
\begin{itemize}
    \item We define Logical Anomaly Classification and present \emph{LogiCls}, which decomposes complex logical constraints into atomic subqueries for interpretable and verifiable reasoning.
    \item We develop a fine-grained data synthesis pipeline with object-grounded reasoning templates and introduce a difficulty-aware resampling strategy that adapts sampling based on model errors and uncertainty.
    \item \emph{LogiCls} with small-scale vision language models outperforms strong baselines in accuracy, improves inference efficiency, and provides transparent reasoning chains that enhance interpretability and robustness.
\end{itemize}

\section{Task Definition and Dataset}
\label{sec: task definition}
\noindent\textbf{Formulation}
We formulate the logical anomaly classification (LAC) task as an end-to-end set prediction problem. 
During training, the model is provided with images labeled as either normal or associated with a single anomaly type from a predefined finite set $ \mathcal{C} = \{c_1, c_2, ..., c_K\} $, where each $c_k$ denotes a distinct anomaly category.
During inference, the model is expected to predict a subset of the extended label space $ \mathcal{C}^+ = \{ \text{normal} \} \cup \mathcal{C} $, such that the output is an element of the power set $ \mathcal{P}(\mathcal{C}^+) $, i.e., a subset of $\mathcal{C}^+$. 
Specifically, if no anomalies are detected, the predicted set is $\{ \text{normal} \}$.
Otherwise, it contains the normal label excluded and one or more anomaly types indicating the detected defects.
Let $ \mathcal{X} $ denote the space of input images. The goal is then to learn a mapping function $ f: \mathcal{X} \rightarrow \mathcal{P}(\mathcal{C}^+) $, which generalizes from training samples with single-label annotations (either normal or a single anomaly type) to real-world scenarios where multiple anomaly types may coexist or none can be present.

\noindent\textbf{MVTec LOCO FC Dataset.}
The MVTec LOCO \cite{bergmann2022mvtechloco} dataset is the most popular industrial logic anomaly dataset and contains five scene categories. 
The training set originally included only normal images, whereas the testing set included both normal and abnormal images.
This dataset is widely used in LAD research, and to our knowledge, the classification task has never been studied on it.
Following the methodology of MVREC \cite{lyu2025mvrec}, we utilize the provided masks to categorize anomalies and resplit the data into \emph{MVTec LOCO FC} (Tab.~\ref{tab: resplit dataset}).
For anomalous samples that are only in the original test set, 80\% are randomly sampled into the new training set and 20\% into the new test set. 
All samples with multiple logical anomalies are manually placed in the new test set.
Samples containing multiple co-occurring logical anomalies are placed exclusively in the test set to evaluate the model's ability to handle complex, real-world scenarios. 
For normal samples, the new dataset is divided in the same way as the original dataset.

\begin{table}[t]
 \caption{Distribution of MVTec LOCO FC dataset.}
    \centering
     \resizebox{.99\linewidth}{!}{    
    \begin{tabular}{lcccccc}
\toprule
Scenario & \multicolumn{2}{c}{\textbf{Training Set}} & \multicolumn{3}{c}{\textbf{Test Set}} \\
\cmidrule(lr){2-3} \cmidrule(lr){4-6}
 & Normal & Single Anomaly  & Normal & Single Anomaly  & Multi Anomaly \\ 
 \midrule
 Breakfast Box & 351& 66&  102&17&0\\
Juice Bottle & 335& 104&  94&26&12\\
 Screw Bag & 360& 90&  122&23&24\\
 Pushpins & 372& 32&  138&48&11\\
  Splicing Connectors & 360& 86&  119&22&0\\
    \bottomrule
    \end{tabular}
    }
    
    \label{tab: resplit dataset}
\end{table}

\begin{figure}[t]
    \centering
    \includegraphics[width=\linewidth]{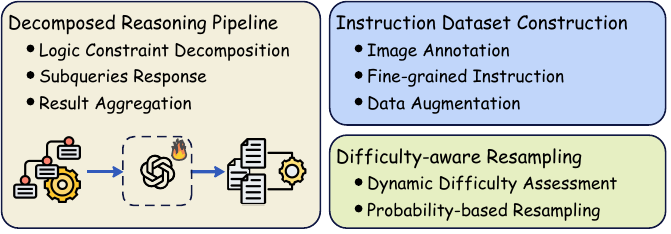}

    \caption{Overview of the proposed \emph{LogiCls} framework. The method first decomposes complex logical constraints into atomic subqueries. Then, an instruction dataset is constructed through image annotation, fine-grained reasoning templates, and data augmentation. During training, a difficulty-aware resampling strategy dynamically adjusts sampling probabilities. Finally, a fine-tuned model performs inference by aggregating subquery outputs for accurate logical anomaly classification.}
   
    \label{fig: method}
\end{figure}

\section{Method}

As illustrated in Fig.~\ref{fig: method}, we propose \emph{LogiCls}, a framework that decomposes complex industrial logical constraints into tractable subqueries and performs anomaly classification by aggregating verifiable subquery answers. 
The framework follows a data-driven strategy: we synthesize instruction data to enhance spatial and counting ability in small-scale VLMs, and we apply difficulty-aware resampling during fine-tuning for robust classification.

\subsection{Logic Constraint Decomposition}
\label{subsec: logic constraint decomposition}
In industrial logical anomaly scenarios, classification requires object recognition together with reasoning over constraints on spatial relations, quantity, and attribute consistency across multiple objects. Direct end-to-end inference to get anomaly types $\hat{y}\in \mathcal{P}(\mathcal{C}^+)$ with a VLM $g_\theta$ on an image $I \in\mathcal{X}$ with corresponding logic constraint text $L$ often yields suboptimal accuracy and limited interpretability:

\begin{equation}
\hat{y} = g_\theta(I, L).
\label{eq:e2e}
\end{equation}

\noindent We therefore decompose $L$ into atomic subqueries $\{q_t\}_{t=1}^{T}$, each targeting a specific feature or violation pattern. For each subquery, we obtain

\begin{equation}
z_t = g_\theta(I,q_t), \quad t=1,\ldots,T,
\label{eq:subquery}
\end{equation}

\noindent and then map the set of answers $\{z_t\}_{t=1}^T$ to final anomaly categories with a scenario-specific aggregator $h$

\begin{equation}
\hat{y} = h(z_1,\ldots,z_T).
\label{eq:h_agg}
\end{equation}

\noindent Each subquery returns a verifiable output such as a numeric value $z_t \in \mathbb{R}$, a boolean value in $\{0,1\}$, or a categorical element from a finite set. The aggregation in Eq.~\ref{eq:h_agg} provides a reasoning trail and a precise mapping to categories $\hat{y}\in\mathcal{P}(\mathcal{C}^+)$.

\subsection{Fine-grained CoT Instruction Data}
Since logical anomalies hinge on spatial and counting constraints and small-scale VLMs underperform in such settings, we build a three-step instruction synthesis pipeline.
\textbf{(1) Image annotation.}
Following LogSAD \cite{zhang2025logsad}, we apply CLIP \cite{radford2021clip} and SAM \cite{kirillov2023sam} for open-vocabulary segmentation \cite{wang2024samclip} on MVTec LOCO FC to obtain masks and location coordinates.
\textbf{(2) Fine-grained instruction.}
For each image $I$ and subquery $q_t$, we define $g_\theta(I,q_t) \rightarrow (c_t, z_t)$, where $c_t$ is a CoT enclosed by \texttt{<think>} and \texttt{</think>} with grounding to bounding boxes, and $z_t$ is the final answer enclosed by \texttt{<answer>} and \texttt{</answer>}. We refine reasoning into direction, distance, size, counting, and other types. Examples are provided in Fig.~\ref{fig:introduction} and Fig.~\ref{fig:case_study}, and the collection forms $D_{\text{cot}}$.
\textbf{(3) Data augmentation.}
To enlarge $D_{\text{cot}}$, we build $D_{\text{aug}}$ with modality specific strategies. For direction and counting, we cut and paste masked objects with Poisson image editing \cite{perez2003poisson} and update the labels. For other types, we add noise and apply style transfer. For text, large language models produce between $10$ and $20$ paraphrases per subquery to increase linguistic diversity.

\subsection{Difficulty-aware Resampling}
\label{subsec: resampling}
We introduce a dynamic resampling scheme that focuses training on challenging subqueries by adjusting sampling probabilities according to an online estimate of difficulty.

\noindent\textbf{Dynamic difficulty assessment.}
At the end of each epoch, the model evaluates the difficulty of every sample in $D_{\text{cot}}$. For sample $i$, we compute a score

\begin{equation}
d_i \;=\; \alpha \cdot \mathbf{1}\!\left[\hat{y}_i \neq y_i\right] \;+\; \beta \cdot \mathrm{Perplexity}(\hat{y}_i),
\label{eq:difficulty}
\end{equation}

\noindent where $\mathbf{1}[\cdot]$ is the indicator function and $\alpha=1$, $\beta=0.2$ in all experiments.



\noindent\textbf{Probability-based resampling.}
Let $\mathcal{Q}_t$ be the set of augmented samples in $\mathcal{D}_{\text{aug}}$ derived from base subquery $q_t$. We assign a sampling probability

\begin{equation}
P_s^{(t)} = 
\frac{\left( \sum_{i \in \mathcal{Q}_t} d_i \right)^{\gamma}}%
{\sum_{l=1}^{T} \left( \sum_{i \in \mathcal{Q}_l} d_i \right)^{\gamma}},
\label{eq:prob}
\end{equation}

\noindent where $\gamma > 0$ controls how strongly difficulty influences sampling. Minibatches are drawn from $\mathcal{D}_{\text{aug}}$ according to $\{P_s^{(t)}\}_{t=1}^{T}$, which increases exposure to hard subqueries while retaining coverage of easier ones.

\section{Experiment}

\subsection{Settings}

\noindent\textbf{Setup.}
As the first study of LAC, we evaluate on MVTec LOCO FC. All inputs use native resolution without resize or crop.
Following prior work \cite{zhang2023icl}, we adopt a unified in-context learning (ICL) format that integrates decomposed subqueries, a single positive training image, the test image, and explicit instructions for both reasoning and output. 
This setup provides a consistent scaffold for the model to infer step-by-step from the exemplars and apply the same reasoning to the test case, as shown in Fig.~\ref{fig:icl}.
Baselines are Gemini-2.5-pro \cite{comanici2025gemini2.5}, GPT4.1-2025-0414\footnote{https://openai.com/index/gpt-4-1/}, GPT-4o \cite{hurst2024gpto}, GLM-4.5V \cite{hong2025glm}, InternVL3-78B-Instruct \cite{zhu2025internvl3} and QwenVL2.5-72B, 7B, 3B-Instruct \cite{bai2025qwenvl2.5}. We also fine-tune 3B and 7B models with \emph{LogiCls}.

\begin{table*}[t]
\caption{The results of MVTec LOCO FC include \textit{binary F1} and multi-classified \textit{macro F1} for logic anomalies. }
    \centering
    \setlength{\tabcolsep}{2pt}
     \resizebox{.99\linewidth}{!}{    
    \begin{tabular}{l|c |cc|cc |cc |cc |cc |cc}
    \toprule
    \multirow{2}{*}{Model}&
        \multirow{2}{*}{Method}&\multicolumn{2}{c}{Breakfast Box}&\multicolumn{2}{|c}{Juice Bottle}& \multicolumn{2}{|c}{Screw Bag}&\multicolumn{2}{|c}{Pushpins}& \multicolumn{2}{|c}{Splicing Connector} &  \multicolumn{2}{|c}{Average} \\
        \cmidrule{3-4}\cmidrule{5-6} \cmidrule{7-8} \cmidrule{9-10} \cmidrule{11-12} \cmidrule{13-14}
        
    &    & binary F1 & macro F1  & binary F1 & macro F1 & binary F1 & macro F1& binary F1 & macro F1& binary F1 & macro F1  & binary F1 & macro F1\\
\midrule
Gemini-2.5-pro \cite{comanici2025gemini2.5} & ICL &93.45 &89.74 &85.56 &39.86 &72.45 &50.87 &40.87& 32.58& 42.39 &48.64&64.23& 49.89\\
GPT4.1-2025-0414 &ICL  &90.75& 83.46 & 81.25 &35.48&63.50 & 42.78 &43.87& 30.65 &43.16 &51.45& 61.98 &45.91\\
GPT-4o \cite{hurst2024gpto} & ICL & 86.67 & 61.82 & 81.25 & 35.48 & 43.48 & 28.44 & 39.71 & 27.67 & 35.90 & 42.86 & 54.45 & 37.39\\
GLM-4.5V$_{no\ think}$ \cite{hong2025glm} & ICL & 55.46 & 65.05 & 57.36&49.74  & 44.67&41.24 & 35.85&38.26 & 53.33& 44.05 & 47.89 & 46.68\\
QwenVL2.5-72B-Instruct \cite{bai2025qwenvl2.5} & ICL & 52.00 & 68.03 & 50.35 & 35.54 & 41.44 & 8.19 & 32.39 & 19.14 & 45.16 & 40.74 & 42.99 & 31.25\\
InternVL3-78B-Instruct \cite{zhu2025internvl3} & ICL & 45.16 & 30.30 & 81.25 & 33.37 & 39.66 & 11.16 & 34.59 & 35.27 & 33.96 & 40.42 & 45.39 & 29.74\\
\midrule
QwenVL2.5-7B-Instruct \cite{bai2025qwenvl2.5} & ICL & 34.67 & 20.15 & 56.60 & 10.48 & 40.35 & 10.38 & 38.02 & 23.80 & 26.09 & 21.30 & 39.03 & 17.45\\
  &\textbf{\emph{LogiCls}}& \textbf{100.0} & \textbf{97.39} & \textbf{90.96} & \textbf{86.98} & \textbf{92.16} & \textbf{88.74} & \textbf{87.80} & \textbf{98.00} & \textbf{95.24} & \textbf{98.41} & \textbf{92.62} & \textbf{94.00}\\
\midrule
QwenVL2.5-3B-Instruct \cite{bai2025qwenvl2.5} & ICL & 21.92 & 3.51 & 13.79 & 0.00 & 40.72 & 8.53 & 34.20 & 13.89 & 27.16 & 3.18 & 28.86 & 6.65\\
  & \textbf{\emph{LogiCls}}& \textbf{100.0} & \textbf{92.16} &\textbf{81.45} & \textbf{75.44} & \textbf{89.52} & \textbf{85.03} & \textbf{79.01} & \textbf{76.89} & \textbf{95.24} & \textbf{94.07} & \textbf{88.09} & \textbf{84.05}\\

    \bottomrule
    
    \end{tabular}
    }

    \label{tab:main result}
\end{table*}

\begin{figure}[t]
    \centering
\begin{quote}
        \colorbox{gray!20!white}{
            \parbox{\linewidth}{
                \raggedright 
                \fontsize{9}{10}\selectfont\ttfamily
             
\textbf{System Prompt:} \\
 - Task: You need to answer questions based on the image uploaded. \\
 - Format: Your response should include a concise reasoning process and an accurate answer, with the reasoning and answer enclosed separately using <think> and </think>, and <answer> and </answer> tags, for example: <think>This is the reasoning process</think><answer>This is the answer</answer>.\\
\textbf{User Prompt:} [SUBQUERY]\\

\textbf{Case:}\\
- Input Image:    [Example Image]\\
- Output:    [Reasoning phase and answer]\\
On the test image, provide a concise reasoning and the corresponding answer.\\
\textbf{Test:}\\
- Input: [Test Image]\\
- Output:
            }
        }
\end{quote}
\caption{Illustration of ICL. The content inside the brackets will be replaced with real samples during the experiments.}
\label{fig:icl}
\end{figure}

\noindent\textbf{Metrics.}
Images may be normal or carry multiple anomaly types, so we report two measures. \emph{Binary F1} collapses all anomaly types into a single anomaly class. \emph{Macro F1} averages the F1 score over all anomaly classes to reflect class-balanced performance.


\subsection{Main Results}
We present \textit{binary F1} and \textit{macro F1} for LAC in  Tab.~\ref{tab:main result}.
First, focusing on the “Average” columns reveals that, without any training, large-scale VLMs generally outperform small-scale VLMs, and closed-source models perform better than open-source models. 
Our model performed the best, demonstrating that small-scale VLMs can be enhanced via our framework.
Comparing the Breakfast Box and Screw Bag scenarios, we find that every method yields higher metrics for the former than for the latter.
In the Breakfast Box, the types of logical anomalies mainly involve the combination of common objects (such as cereal, starfruit, oranges, etc.), while in the Screw Bag, the anomalies are all related to counting similar objects (such as screws, nuts, and washers of different lengths).
This indicates that different types of logical anomalies present varying levels of difficulty for VLMs. 
Counting similar objects is more challenging than identifying the combination anomalies of common objects.

\begin{table}[t]
\caption{Ablation study on components of our framework.}
\begin{center}

    \resizebox{.99\linewidth}{!}{    
    \begin{tabular}{ccccc|cc}
    \toprule
         Decomp.& Reason. & Bbox. & Resample. &SFT& Avg.binary F1 & Avg.macro F1 \\
         \midrule
         \checkmark  & \textcolor{gray!60}{\ding{55}} & \textcolor{gray!60}{\ding{55}}  & \textcolor{gray!60}{\ding{55}}  &  \textcolor{gray!60}{\ding{55}}  &40.18 & 27.24 \\
         \checkmark  &  \textcolor{gray!60}{\ding{55}} & \textcolor{gray!60}{\ding{55}}  &  \textcolor{gray!60}{\ding{55}} & \checkmark & 45.34&24.20 \\
         \checkmark & \checkmark &  \textcolor{gray!60}{\ding{55}} & \textcolor{gray!60}{\ding{55}}  & \checkmark & 55.77&39.31\\
        \checkmark & \checkmark & \checkmark&  \textcolor{gray!60}{\ding{55}}  & \checkmark & 72.37&51.62 \\
          \checkmark & \checkmark & \checkmark& \checkmark& \checkmark & \textbf{92.62}& \textbf{94.00}\\
    \bottomrule
    \end{tabular}
    }

    \label{tab: ablation study}
\end{center}
\end{table}

\subsection{Ablation Study}
We conduct ablation experiments on the MVTec LOCO FC dataset using the QwenVL2.5-7B-Instruct model \cite{bai2025qwenvl2.5}. The results, summarized in Tab.~\ref{tab: ablation study}, reveal the contribution of each component in our proposed method.

\noindent\textbf{Constraint Decomposition.}
We first decompose the original logical constraints into subqueries and directly prompt the model. This experiment establishes the baseline for subsequent enhancements.

\noindent\textbf{Supervised Fine-tuning.}
Next, we fine-tune the model using the subqueries, corresponding images, and their annotated answers. This step evaluates the effectiveness of fine-tuning beyond simple prompting. While it yields modest improvements, the gains are limited.

\noindent\textbf{Distilled CoT.}
Building on fine-tuning, we incorporate reasoning CoTs into the model outputs. The CoTs are distilled from QwenVL2.5-72B-Instruct \cite{bai2025qwenvl2.5} but exclude bounding box information. This addition leads to clear improvements, particularly in \textit{macro F1} ($39.31, +15.11$), highlighting the benefit of explicit intermediate reasoning.

\noindent\textbf{Fine-grained CoT.}
We further enhance the reasoning chain by embedding the spatial coordinates of the target objects. 
This spatial grounding produces significant performance gains ($72.37/51.62, +16.60/$$+12.31$), confirming the necessity of aligning relational constraints with explicit object locations to ensure logical consistency.

\noindent\textbf{Difficulty-aware Resampling.}
Finally, based on the fine-grained CoT dataset, we further improve the two indicators to $92.62/94.00$ ($+20.25/$$+42.38$) by using image augmentation and difficulty-aware resampling (\S \ref{subsec: resampling}), reaching the optimum. This indicates that the long tail and the bias in the distribution of sample difficulty are effectively mitigated and that the model benefits from being balanced across categories.

In conclusion, the combination of spatial-aware reasoning and data balancing strategies constitutes the primary driver of the model's robust logical reasoning capabilities.


\begin{figure*}[t]
    \centering
    \begin{minipage}[t]{0.45\textwidth}
        \centering
        \includegraphics[width=\linewidth]{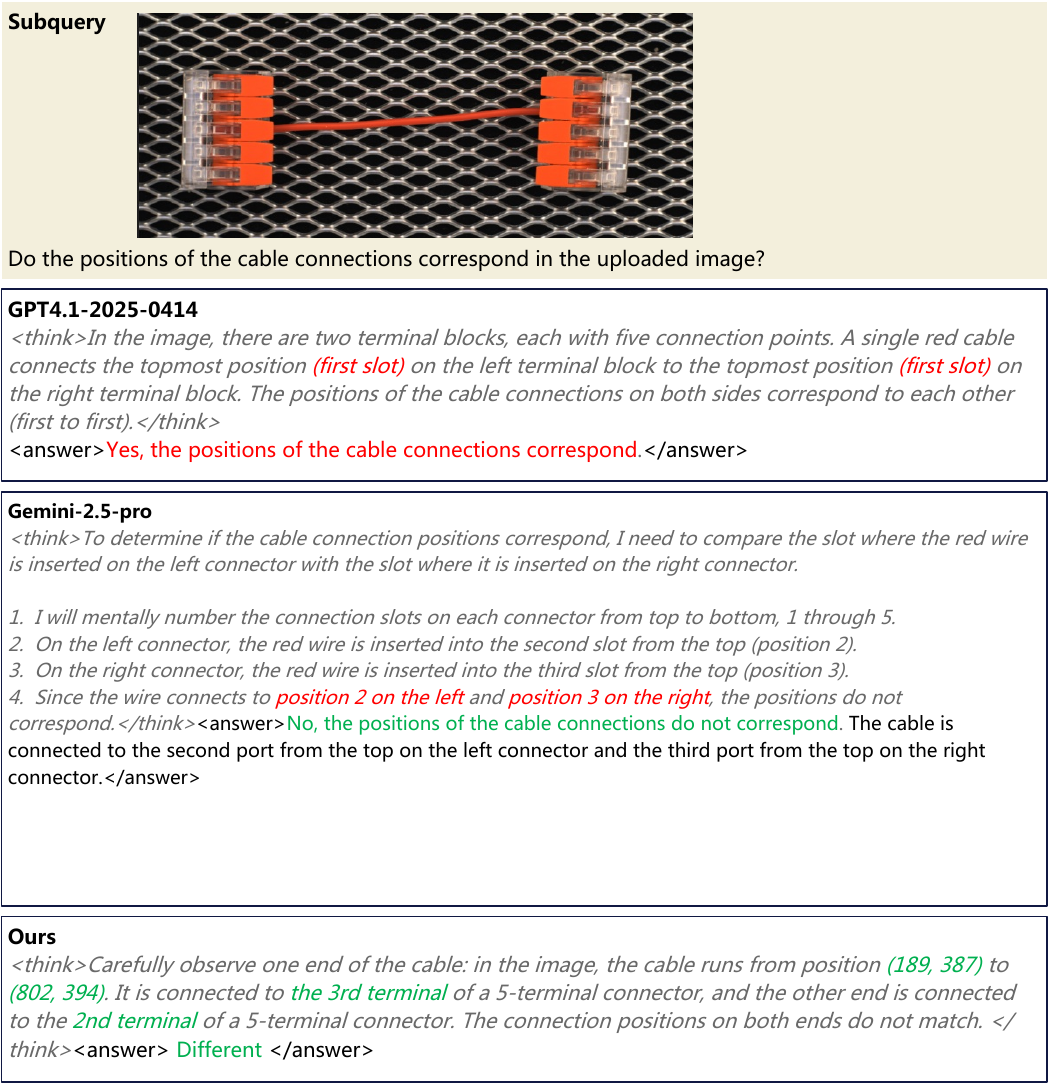}
        \footnotesize{(a)}
        \label{fig:sub1} 
    \end{minipage}
    \hfill
    \begin{minipage}[t]{0.45\textwidth}
        \centering
        \includegraphics[width=\linewidth]{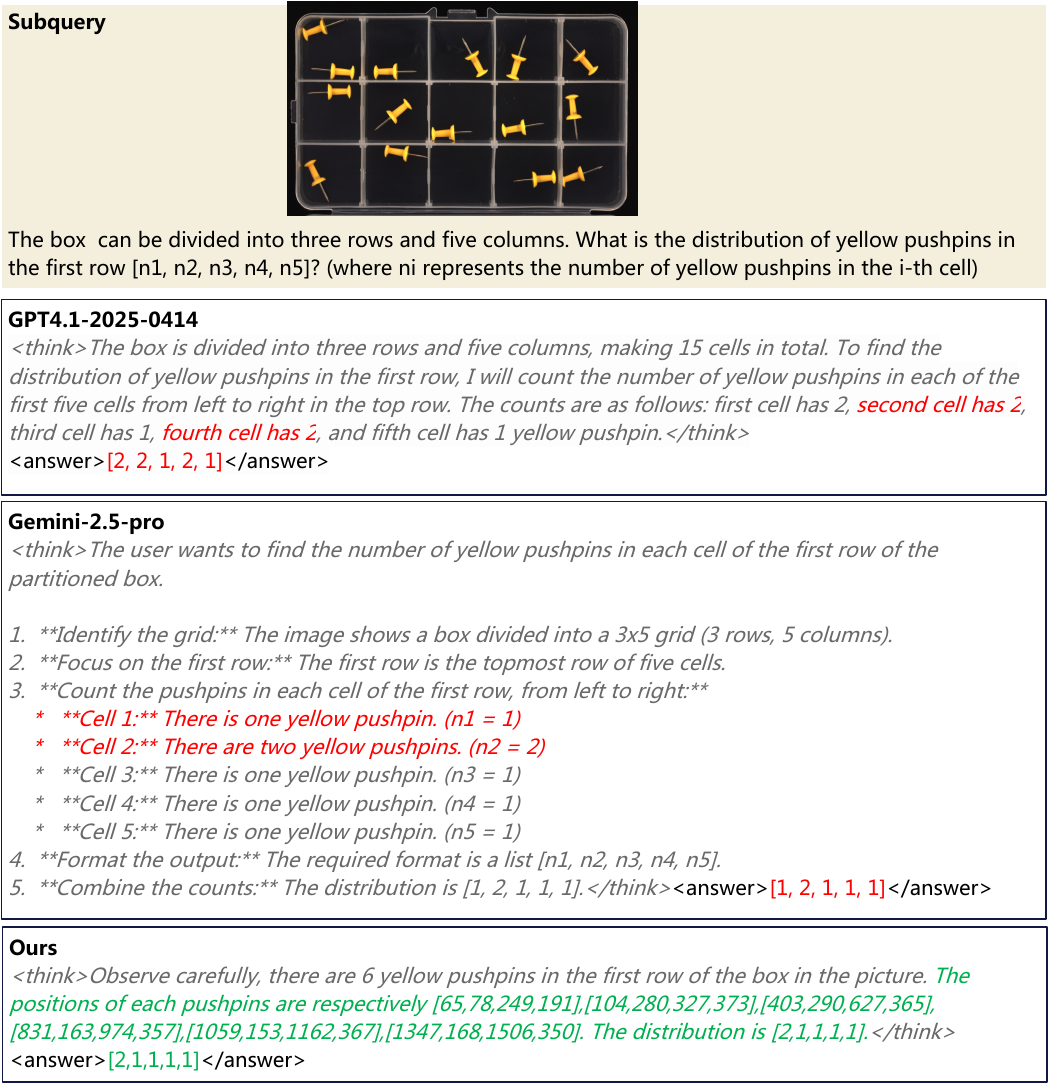}
        \footnotesize{(b)}
        \label{fig:sub2}
    \end{minipage}
    \vfill
    \begin{minipage}[t]{0.45\textwidth}
        \centering
        \includegraphics[width=\linewidth]{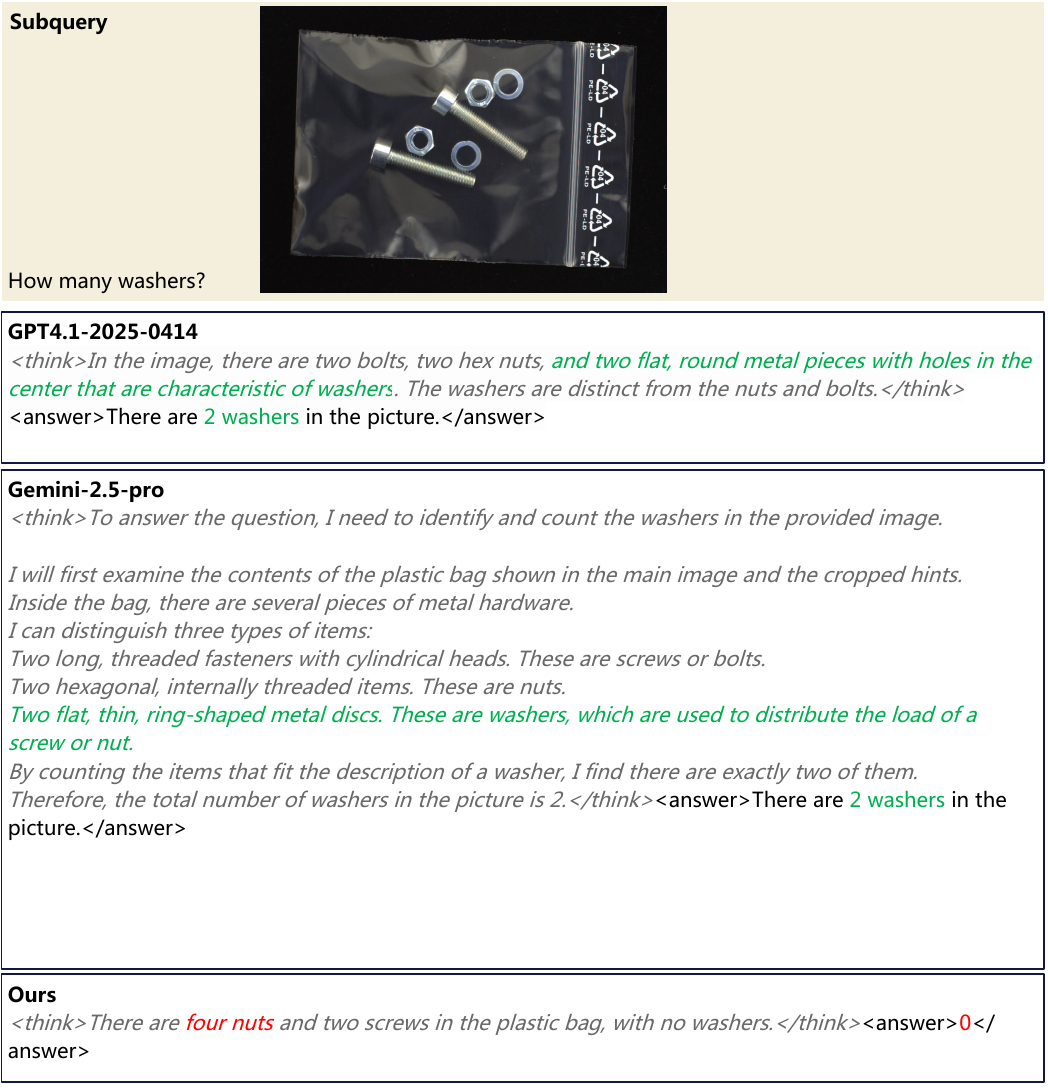}
       \footnotesize{(c)}
        \label{fig:sub3}
    \end{minipage}
    \hfill
    \begin{minipage}[t]{0.45\textwidth}
        \centering
        \includegraphics[width=\linewidth]{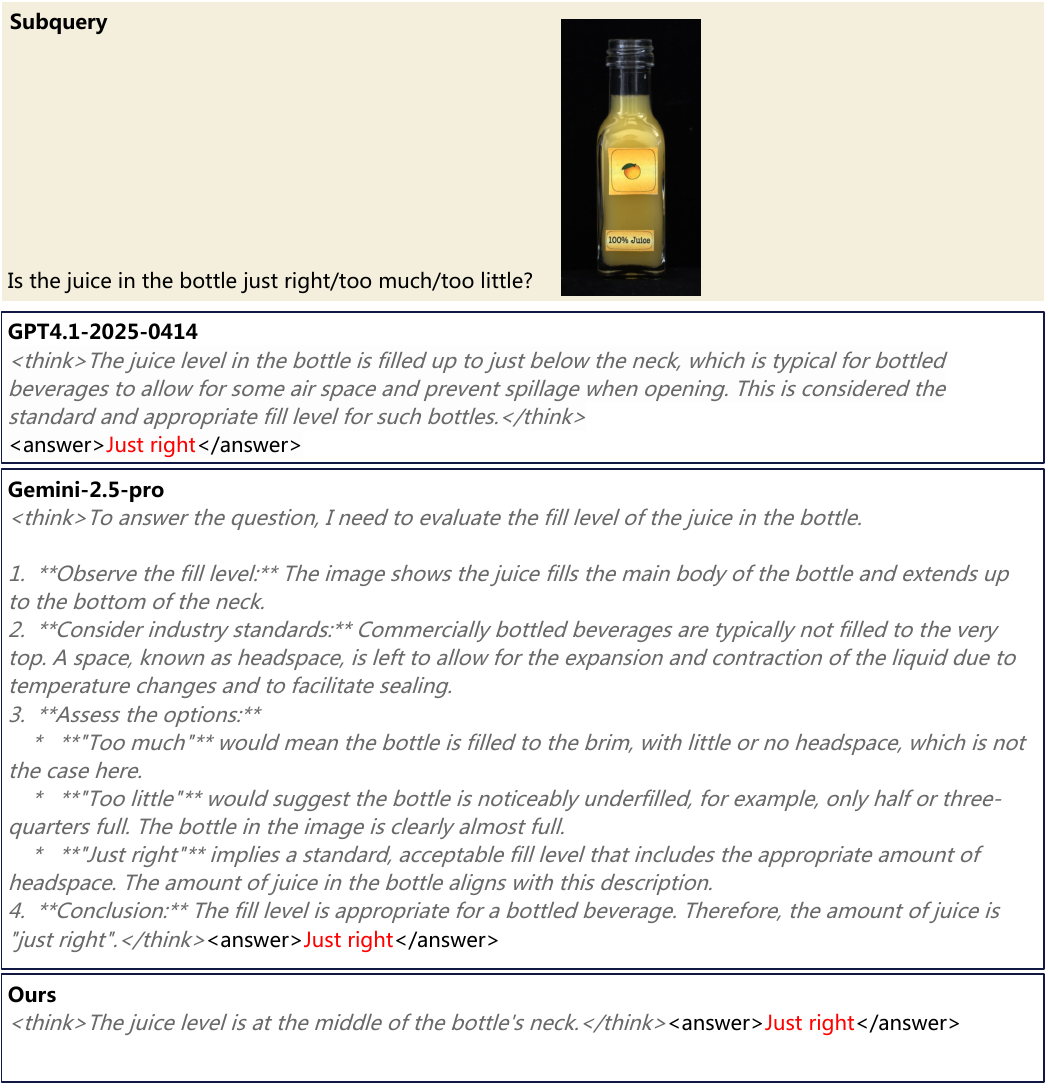}
        \footnotesize{(d)}
        \label{fig:sub4}
    \end{minipage}
    \caption{A comparison with VLMs on the decomposed subqueries. Correct answers are in green and incorrect answers are in red.}
    \label{fig:case_study} 
\end{figure*}

\subsection{Case Study}

We further conduct a fine-grained qualitative assessment to understand the specific benefits and limitations of our coordinate-aware reasoning approach. By analyzing the intermediate reasoning traces of our model alongside GPT4.1-2025-0414 and Gemini-2.5-pro, we highlight differences in spatial grounding capabilities. The comparison covers four distinct visual reasoning tasks.

\noindent\textbf{Spatial Reasoning.}
In Fig.~\ref{fig:case_study}a, the cable connection task reveals significant disparities in spatial logic between the models. GPT4.1-2025-0414  fails the task by incorrectly claiming the red cable connects the first slot on both sides, leading to a false ``Yes'' answer. While Gemini-2.5-pro correctly identifies that the positions do not correspond, it misidentifies the specific slots.
In contrast, our model accurately grounds the connection through precise coordinates, and correctly identifies the cable as being connected to the 3rd terminal on one end and the 2nd terminal on the other. This demonstrates that coordinate-level fine-grained CoT is essential for resolving complex spatial relationships that generic reasoning traces often misinterpret.

\noindent\textbf{Dense Grid Counting.}
As illustrated in Fig.~\ref{fig:case_study}b, the task requires counting within specific grid partitions. 
The reasoning traces of baseline models, such as GPT4.1-2025-0414 and Gemini-2.5-pro, reveal counting discrepancies within individual cells. 
In contrast, our model explicitly generates spatial coordinates for each pushpin during the reasoning phase, allowing it to derive accurate cell-wise counts and the exact distribution. 
This mechanism results in correct cell-level counts and the exact final distribution, suggesting that spatial-aware fine-grained CoT is critical for mitigating counting hallucinations.

\noindent\textbf{Object Identification.}
As illustrated in Fig.~\ref{fig:case_study}c, both GPT4.1-2025-0414 and Gemini-2.5-pro predict the correct answer, successfully distinguishing between visually similar objects (washers and nuts) within their reasoning traces. 
In contrast, our model failed to identify the washers, misclassifying them as nuts. This suggests that our current method does not significantly improve fine-grained discrimination for confusingly similar targets, resulting in performance that falls short of the baseline models in this specific scenario.

\noindent\textbf{Relative Scale Estimation.} As shown in Fig.~\ref{fig:case_study}d, the fill-level assessment task, which requires precise relative scale estimation, reveals a collective limitation in current models. Both proprietary baselines and our model incorrectly conclude that the liquid level is ``Just right'', failing to perceive the subtle anomaly. 
This failure suggests that subtle relative concepts, as opposed to absolute ones, are inherently ambiguous, highlighting the limitations of current ICL and fine-tuning methodologies in addressing such nuances.

Qualitative analysis demonstrates that \emph{LogiCls} significantly enhances spatial grounding and dense counting accuracy compared to proprietary baselines, effectively mitigating hallucinations through precise coordinate-level CoT. 
However, the approach shows limited improvement in fine-grained object discrimination and subtle relative scale estimation, where it shares common pitfalls with existing models.

\section{Conclusion}
In this paper, we introduce the LAC task and propose a multi-model framework \emph{LogiCls} to fine-tune small-scale VLMs.
Our approach first takes a data-centric approach by using logic composition, constructing fine-grained CoT data, and performing data augmentation. 
Subsequently, we employ a difficulty-aware resampling strategy to improve the model's understanding of industrial scenarios.
However, our approach struggles with visually similar categories and subtle scale anomalies. Future work will integrate hierarchical geometric priors and global-to-local attention to improve structural consistency and fine-grained scale sensitivity.

{\small
\bibliographystyle{IEEEbib}
\bibliography{refs}
}

\end{document}